\documentclass[11pt]{article}

\PassOptionsToPackage{table}{xcolor}

\usepackage[preprint]{acl}

\usepackage{times}
\usepackage{latexsym}

\usepackage[T1]{fontenc}

\usepackage[utf8]{inputenc}

\usepackage{microtype}

\usepackage{inconsolata}

\usepackage{graphicx}
\usepackage{amsmath}
\usepackage{amssymb}
\usepackage{booktabs}
\usepackage{multirow}
\usepackage{array}
\usepackage{tabularx}
\usepackage{subcaption}

\definecolor{badbg}{HTML}{FDECEA}
\definecolor{badborder}{HTML}{C62828}
\definecolor{goodbg}{HTML}{E8F5E9}
\definecolor{goodborder}{HTML}{2E7D32}
\definecolor{neutralbg}{HTML}{F5F5F5}
\definecolor{neutralborder}{HTML}{616161}

\title{ARBOR: Online Process Rewards via a Reusable Rubric Buffer \\for Search Agents}

\author{Zheng Liu$^{1,*}$, \quad
Longxiang Zhang$^{2}$, \quad
Xintong Wang$^{2}$, \quad
Zhiang Xu$^{2}$, \quad
Shaoxiong Zhan$^{1}$, \\
\textbf{Xin Shan$^{3}$, \quad
Wen Huang$^{1}$, \quad
Tao Dai$^{4,\dagger}$, \quad
Shu-Tao Xia$^{1}$, \quad
Chengfu Huo$^{2}$, \quad
Liang Ding$^{2,\dagger}$} \\
$^{1}$ Tsinghua University \quad 
$^{2}$ Alibaba Group \quad
$^{3}$ Peking University \quad
$^{4}$ Shenzhen University\\
\texttt{liu-z24@mails.tsinghua.edu.cn}, \quad
\texttt{\{daitao.edu, liangding.liam\}@gmail.com}, \quad
}

\begin{document}
\maketitle
\renewcommand{\thefootnote}{\fnsymbol{footnote}}
\footnotetext[1]{Work done during an internship at Alibaba.}
\footnotetext[2]{Corresponding authors.}
\renewcommand{\thefootnote}{\arabic{footnote}}
\begin{abstract}
LLM-based search agents are trained predominantly with outcome-only reward, leaving the search process itself unsupervised. This signal degenerates on outcome-homogeneous groups where all sampled trajectories share the same correctness, yielding zero within-group advantage and no gradient. Existing process supervision either trains a costly verifier or generates per-query rubrics that are inconsistent across queries and discarded after one use. We propose \textbf{ARBOR} (\textbf{A}daptive \textbf{R}ubric \textbf{B}uffer for \textbf{O}nline \textbf{R}eward), a reusable process-reward framework that maintains a rubric memory shared across queries. Query-local drafts induced from contrastive trajectories are admitted, consolidated into cross-query common rubrics, and retired as the policy evolves. A small active subset of common rubrics scores trajectories via sparse pairwise judging, and the resulting scores are added to the base reward, providing process-level gradient even when outcome reward is uniform. ARBOR consistently outperforms GRPO and DAPO baselines on four multi-hop QA benchmarks, raising average LLM-judge accuracy by up to \textbf{4.2} points and converting up to \textbf{42\%} of otherwise-zero-gradient training groups into informative ones.\footnote{We will release our code and models after the review process.}
\end{abstract}

\section{Introduction}
\label{sec:introduction}

\begin{figure}[t]
    \centering
    \includegraphics[width=\linewidth]{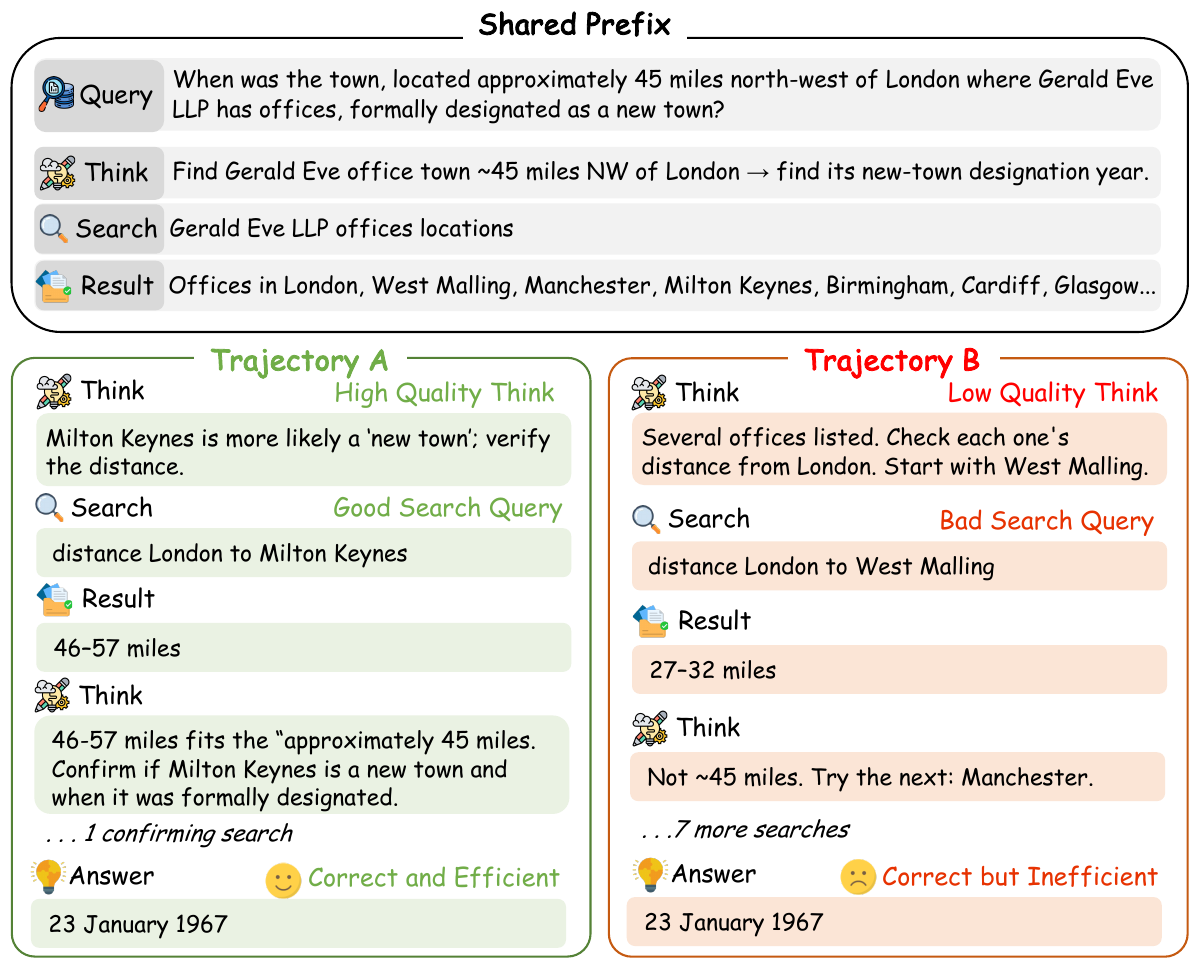}
    \caption{\textbf{Process quality divergence under identical outcomes.} Two trajectories from the same query reach the same answer yet differ markedly in search efficiency.}
    \label{fig:fig1}
  \end{figure}

LLM-based agents that interact with external environments under reasoning-and-acting paradigms such as ReAct~\citep{yao2023react} have become a standard approach to complex tasks. A representative case is the search agent~\citep{press-etal-2023-measuring,xi2025deepsearchagents}, which answers multi-hop questions requiring external knowledge by iteratively rewriting queries, retrieving evidence, filtering observations, and integrating them into a final answer. This interaction pattern lets search agents substantially outperform LLMs that answer directly on multi-hop QA and other complex information-retrieval tasks. Recent systems such as Search-R1~\citep{jin2025searchr1} and R1-Searcher~\citep{song2025r1searcher}, together with other search-agent RL studies~\citep{li-etal-2025-search,jiang-etal-2025-s3}, further show that RL is an effective way to push the capability ceiling of search agents and has become the dominant paradigm for their training. Within this paradigm, the RL stage relies almost exclusively on outcome-only reward, using final-answer correctness as the reward signal together with a format penalty~\citep{shao2024deepseekmath,yu2025dapo}, and provides no supervision on the search process itself.

Trajectories sampled from the same query can follow very different search paths even when they arrive at the same outcome, as illustrated in Figure~\ref{fig:fig1}: one may reason carefully through targeted retrieval while another stumbles onto the answer after redundant detours, but final-answer correctness assigns them identical reward. Under group-relative objectives such as GRPO~\citep{shao2024deepseekmath}, identical within-group reward produces zero relative advantage and no policy gradient, so process differences that could inform better search behavior contribute nothing to training. Such outcome-homogeneous groups are far from rare in search-agent RL training (see Section~\ref{sec:no-gradient-rescue}), making them a major bottleneck for outcome-only reward.

Adding process-level supervision is a natural response, yet existing routes do not fit search agents cleanly. Training a process reward model (PRM) requires rollout-based annotation or value estimation over intermediate reasoning states~\citep{lightman2024verify,wang-etal-2024-math,luo2024omegaprm,cui2025prime}; in search agents, this can require rolling back from intermediate states and invoking search APIs at prohibitive cost, while forcing the inherently qualitative nature of search behavior into discrete step-correctness labels. Query-specific LLM-generated rubrics, exemplified by Rubrics-as-Rewards~\citep{gunjal2025rar}, sidestep verifier training but produce inconsistent criteria across queries and are discarded after one use, so they cannot stably reflect process regularities across queries or evolve with the policy.

These limitations point to three properties that a process reward suited to search-agent RL should possess. First, it should supervise the search process itself, supplying learning signal for the process quality that outcome-only reward overlooks; this property matters most on outcome-homogeneous groups, where process supervision is the only available within-group signal. 
Second, the process criteria should be general, cross-query reusable standards, rather than query-specific rubrics that may conflict across queries.
Third, the effectiveness of process criteria decays as the policy's behavior distribution evolves, so the criteria themselves must be updated continuously rather than fixed once.

\begin{figure*}[t]
    \centering
    \includegraphics[width=\linewidth]{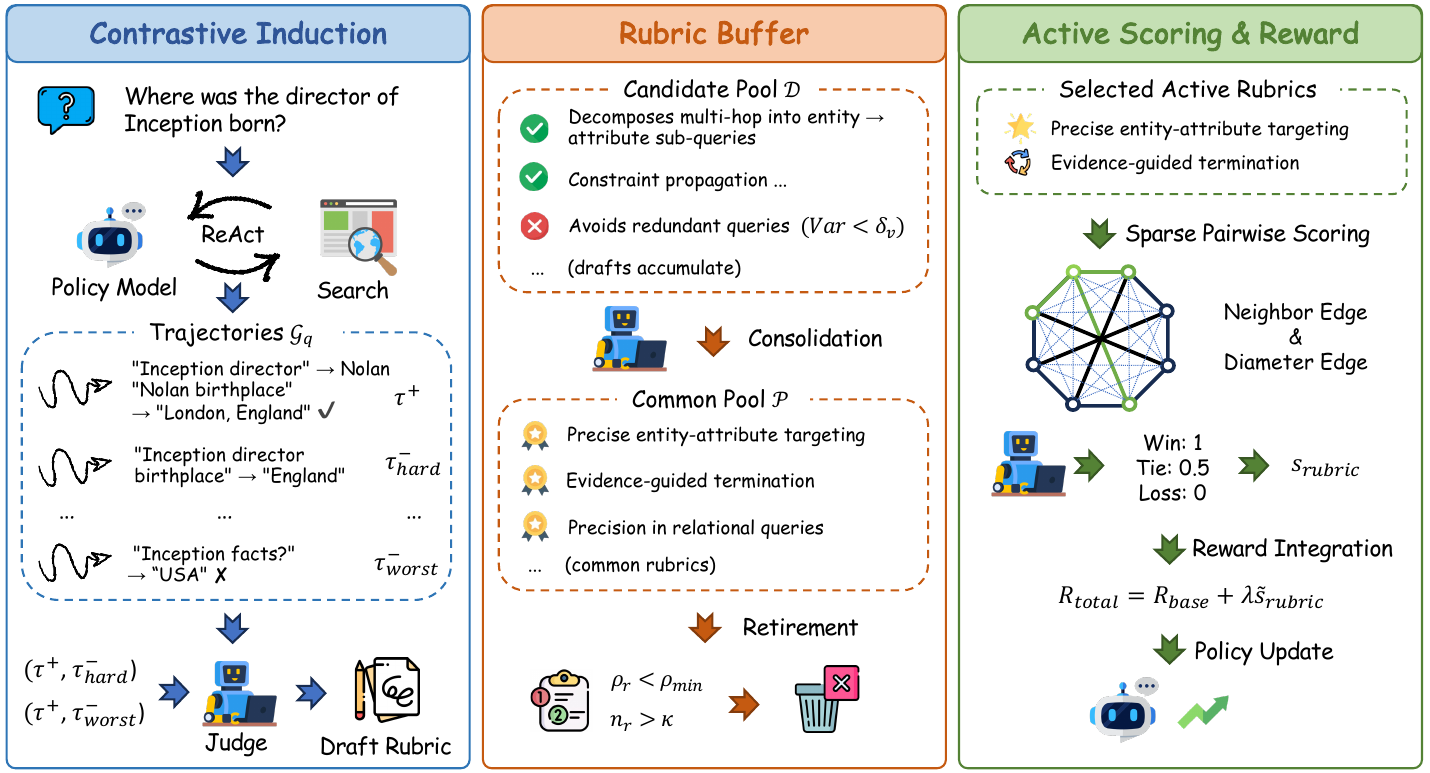}
    \caption{\textbf{Overview of ARBOR.} (a) Contrastive induction extracts query-local draft rubrics from trajectories within a query-group. (b) The rubric buffer $\mathcal{M}$ admits drafts into a candidate pool $\mathcal{D}$, consolidates them into a common pool $\mathcal{P}$, and retires stale rubrics, forming an online admission--consolidation--retirement lifecycle. (c) At each step, two active common rubrics are selected and used to score trajectories via sparse pairwise scoring, and the centered rubric scores are added to the base reward.}
    \label{fig:overview}
\end{figure*}

We propose \textbf{ARBOR} (\textbf{A}daptive \textbf{R}ubric \textbf{B}uffer for \textbf{O}nline \textbf{R}eward), a reusable process-reward framework for search-agent RL training. The core component is a rubric memory consisting of a candidate pool, which holds query-local drafts induced from contrastive trajectories within a query-group, and a common pool, which stores rubrics consolidated into reusable cross-query process criteria. An online lifecycle of admission, consolidation, and retirement consolidates candidate drafts into common rubrics and retires stale ones, so that the common pool provides a unified criterion across queries and evolves with the policy's behavior distribution. Reward shaping invokes only a small active subset of common rubrics. Trajectories within a query-group are scored pairwise under each active rubric, and the resulting scores are added to the base reward. The reusable common pool remains effective even on outcome-homogeneous groups, where it still yields within-group process discrimination that outcome-only reward cannot. Figure~\ref{fig:overview} shows the overall framework.

Our contributions are as follows: 
(1) we propose ARBOR, a reusable process-reward framework that provides within-group process supervision even when outcome-only reward yields zero gradient;
(2) we design a rubric memory with an online admission, consolidation, and retirement lifecycle that maintains consistent cross-query process criteria and evolves with the policy; and 
(3) ARBOR consistently outperforms GRPO and DAPO across three Qwen3 scales on four multi-hop QA benchmarks, improving average LLM-judge accuracy by up to \textbf{4.2} points and converting up to \textbf{42\%} of outcome-homogeneous groups into ones with nonzero reward variance.

\section{Related Work}

\subsection{RL Reward Design for Search Agents}

Reward design for RL-trained search agents has evolved along three lines. The dominant paradigm relies on outcome-only reward, using final-answer correctness or F1 plus format penalties, and leaves the search process entirely unsupervised. Search-R1~\citep{jin2025searchr1}, R1-Searcher~\citep{song2025r1searcher}, and Search-o1~\citep{li-etal-2025-search} are representative of this approach. A second line augments outcome reward with task-specific process heuristics such as information gain, path coverage, or retrieval cost, as in StepSearch~\citep{zheng-etal-2025-stepsearch}, Search-P1~\citep{xia2026searchp1}, SIGHT~\citep{zhong2026sight}, InfoFlow~\citep{luo2025infoflow}, and TIPS~\citep{xie2026tips}, but these metrics cannot capture qualitative aspects of search strategy. A third line trains a process reward model (PRM) to provide step-level feedback, with supervision constructed in different ways: PRM800K~\citep{lightman2024verify} uses human annotation, Math-Shepherd~\citep{wang-etal-2024-math} and OmegaPRM~\citep{luo2024omegaprm} use rollout-value estimation, and PRIME~\citep{cui2025prime} infers from outcomes. In the search-agent setting, representative attempts include PPR~\citep{xu2025ppr} with pre-defined principles and a category-aware PRM, ReasonRAG~\citep{zhang2025reasonrag} with MCTS-constructed step-level annotations followed by process-supervised DPO, and LeTS~\citep{zhang-etal-2025-lets} with a mixture of stepwise process reward and outcome reward.

Existing work either ignores the quality of the search process entirely, reduces it to quantifiable but semantically shallow domain metrics, or relies on a separately trained verifier. Against this background, ARBOR provides process-level feedback that continues to discriminate within a query-group even when its outcomes are homogeneous, restoring a learning signal precisely where outcome-only reward fails.

\subsection{Rubric-Based Reward Signals}

While rubric-conditioned evaluators such as Prometheus~\citep{kim2024prometheus} and LLM-Rubric~\citep{hashemi-etal-2024-llm} use predefined rubrics to structure LLM-as-a-judge assessment, recent work applies rubrics directly as RL reward signals. 
Most methods generate rubrics per query without cross-query sharing, which may cause conflicts across queries. Rubrics-as-Rewards~\citep{gunjal2025rar} uses static query-specific checklists as on-policy rewards; similar per-query approaches include~\citep{wang2025infimedorbit,he2025rifl,zhou2025ruscarl}. Several systems further evolve rubrics or their generators during training~\citep{xu2026rubricarm,sheng2026rlcer,shao2025drtulu}, addressing staleness but still without sharing across queries.
Auto-Rubric~\citep{xie2025autorubric}, AdaRubric~\citep{ding2025adarubric}, and OpenRS~\citep{jia2026openrs} achieve reusability across instances through offline rubric generation, but do not co-evolve with the policy during RL training.

ARBOR combines cross-query reusability with online adaptation: a persistent rubric buffer consolidates query-local drafts into shared common rubrics and continuously retires outdated ones, keeping process standards consistent across queries and aligned with the evolving policy.

\section{Method}

\subsection{Overview and Problem Setup}
\label{sec:overview}

We consider RL training for search agents. Given a query $q$, the policy produces a trajectory $\tau$ through multiple rounds of think, search, and observe interactions and finally outputs an answer. At each training step, we sample $K$ trajectories from the same query to form a query-group $\mathcal{G}_q = \{\tau_1, \ldots, \tau_K\}$, where trajectories share the same query and environment and diverge only through policy sampling.

During training, ARBOR maintains a single rubric memory $\mathcal{M}$ shared across all queries. At every step, ARBOR uses a small number of currently active natural-language process rubrics from $\mathcal{M}$ to score the trajectories in $\mathcal{G}_q$ along process dimensions, and the resulting score is additively combined with the existing RL reward as an auxiliary process-level signal for the policy optimizer. $\mathcal{M}$ itself evolves throughout training under an online admission, consolidation, and retirement lifecycle, with its contents updated in step with the policy behavior distribution.

\subsection{Contrastive Local Rubric Induction}
\label{sec:induction}

ARBOR induces query-local draft rubrics from each query-group $\mathcal{G}_q$ through contrastive induction. Let $F_1(\tau)$ denote the token-level F1 of trajectory $\tau$ against the gold answer. We select the highest-scoring trajectory as a positive anchor, the lowest-scoring trajectory as a worst-case negative, and the strongest remaining trajectory as a hard negative:
\begin{equation}
\label{eq:contrast-pairs}
\begin{aligned}
\tau^{+} &= \arg\max_{\tau \in \mathcal{G}_q} F_1(\tau),\\
\tau^{-}_{\text{worst}} &= \arg\min_{\tau \in \mathcal{G}_q} F_1(\tau),\\
\tau^{-}_{\text{hard}} &= \arg\max_{\tau \in \mathcal{G}_q \setminus \{\tau^{+}\}} F_1(\tau).
\end{aligned}
\end{equation}
These anchors define two complementary contrasts. The pair $(\tau^{+}, \tau^{-}_{\text{worst}})$ exposes large-scale success-failure differences and surfaces critical process deviations, whereas $(\tau^{+}, \tau^{-}_{\text{hard}})$ provides a finer-grained comparison between trajectories of similar correctness and isolates more subtle process differences. Both pairs are provided jointly to an external LLM, which summarizes a small set of natural-language process rubrics as query-local drafts. This design captures both coarse and fine process distinctions while reducing induction from exhaustive $O(K^2)$ pairing to constant cost.

The induction prompt is restricted to search behavior rather than answer content. The LLM is instructed to identify process behaviors that causally separate successful trajectories from failed ones, focusing on search strategy and reasoning process, such as query formulation, evidence use, and stopping judgments. Each induced rubric specifies both what a high-scoring response does and what a low-scoring response does along the dimension.
The full induction prompt is provided in Appendix~\ref{sec:app-prompt-induction}.

All-correct and all-wrong groups skip
the induction stage, as they provide no correctness contrast from which to infer new rubrics.
Common rubrics already saved in $\mathcal{M}$, however, can still score these groups under Section~\ref{sec:reward}, providing within-group discrimination when the outcome signal collapses. Corner cases are detailed in Appendix~\ref{sec:app-induction}.

The resulting drafts enter the memory pipeline of Section~\ref{sec:memory} and can affect future training only after surviving admission and consolidation.

\subsection{Rubric Memory and Lifecycle}
\label{sec:memory}

ARBOR maintains a single rubric memory $\mathcal{M} = (\mathcal{D}, \mathcal{P})$ shared across all queries. The candidate pool $\mathcal{D}$ temporarily stores the query-local drafts induced in Section~\ref{sec:induction}, awaiting abstraction into reusable standards. The common rubric pool $\mathcal{P}$ stores rubrics that have been distilled from $\mathcal{D}$ and serves as the signal source for reward shaping in Section~\ref{sec:reward}. Throughout training, $\mathcal{M}$ evolves through three mechanisms, namely admission, consolidation, and retirement, keeping its contents updated in step with the policy behavior distribution.

\paragraph{Admission.}
Each draft rubric scores the trajectories in its source group via the pairwise judging procedure of Section~\ref{sec:reward}. Let $s_i^r$ denote the score of trajectory $\tau_i$ under draft $r$.
A draft is admitted to $\mathcal{D}$ only if its scores satisfy two conditions:
\begin{equation}
\label{eq:admission}
\begin{aligned}
\mathrm{Var}_{i \in \mathcal{G}_q}(s_i^r) &\geq \delta_v, \\
\mathrm{Pearson}\bigl(\{s_i^r\},\, \{F_1(\tau_i)\}\bigr) &\geq \rho_{\min}.
\end{aligned}
\end{equation}
The variance condition rules out drafts that fail to discriminate on the samples they were induced from. The correlation condition requires that rubric scores are aligned with outcome correctness, rejecting drafts that penalize correct behavior. Drafts failing either condition are discarded immediately.

\paragraph{Consolidation.}
Once the number of drafts in $\mathcal{D}$ reaches a fixed threshold, ARBOR sends all candidate drafts together with the existing common rubrics in $\mathcal{P}$ to an external LLM. The LLM is instructed to identify recurring process patterns across the candidates and abstract them into cross-query general standards. Existing rubrics in $\mathcal{P}$ are provided as context so that the LLM only produces standards covering new dimensions. Each output rubric is further deduplicated against $\mathcal{P}$ by sentence-embedding similarity as a safeguard. After consolidation, $\mathcal{D}$ is cleared. The full consolidation prompt is provided in Appendix~\ref{sec:app-prompt-consolidation}.

\paragraph{Retirement.}
Common rubrics are not assumed to remain valid for the entire training run. As the policy evolves, a rubric that once discriminated good from bad may lose its power because the policy has uniformly mastered or failed the corresponding behavior. ARBOR tracks two long-term signals for each common rubric $r \in \mathcal{P}$, corresponding to the same two dimensions checked at admission stage but accumulated over the rubric's entire active lifetime. 
The first is a consecutive low-variance count $n_r$, the number of consecutive activations on which the within-group score variance under $r$ falls below $\delta_v$, measuring whether $r$ still discriminates among trajectories. The second is a cumulative Pearson correlation $\rho_r$ between the per-trajectory scores under $r$ and the trajectory F1 across all activations of $r$, measuring whether $r$ aligns with correct behavior. Either $n_r$ exceeding its tolerance or $\rho_r < \rho_{\min}$ triggers removal from $\mathcal{P}$.

In addition, when consolidation attempts to write a new common rubric into a $\mathcal{P}$ that has already reached its capacity limit, the entry with the lowest cumulative within-group variance is replaced, provided its tracked statistics have matured over a sufficient number of activations. This prevents newly admitted rubrics from being evicted before their quality signals stabilize.

\subsection{Process Scoring and Reward Shaping}
\label{sec:reward}

At each training step, ARBOR selects a small active subset of common rubrics from $\mathcal{P}$, scores the trajectories in $\mathcal{G}_q$ under those rubrics via pairwise process comparison, and integrates the resulting scores into the reward after within-group centering. Because common rubrics are cross-query process standards independent of any specific query, they can discriminate among trajectories even in outcome-homogeneous groups where the base reward is identical for all samples, sustaining process-level supervision when the outcome signal collapses.

\paragraph{Active selection.}
At each step, ARBOR activates exactly two common rubrics from $\mathcal{P}$. One slot is held by the rubric with the highest cumulative correlation $\rho_r$, serving as the primary process signal; the other rotates among the next strongest candidates by a least-recently-used policy. Activating only two rubrics avoids the linear blowup of judge cost and the within-group signal dilution that would result from scoring under the full pool. Rotation ensures that non-top rubrics still accumulate $\rho_r$ and $n_r$, keeping the retirement mechanism of Section~\ref{sec:memory} operational. When $\mathcal{P}$ is empty, the step falls back to the base reward, so the cold-start phase introduces no unverified rubric signal.

\paragraph{Group-stage process scoring.}
Under each active common rubric, ARBOR performs within-group pairwise comparisons among the trajectories in $\mathcal{G}_q$ via an external LLM judge that returns \textit{win}, \textit{tie}, or \textit{loss}. A full round-robin would require $O(|\mathcal{G}_q|^2)$ judge calls, which is prohibitively expensive. ARBOR therefore sorts the trajectories by base reward and builds a sparse connected graph with $O(|\mathcal{G}_q|)$ edges using two kinds of edges. As illustrated in Figure~\ref{fig:overview}, neighbor edges link adjacent trajectories in the sorted order, providing fine-grained local comparisons between similarly ranked trajectories; diameter edges connect distant trajectories, providing high-contrast pairs for more decisive judgments. Each pairwise call randomizes the presentation order of the two trajectories, eliminating positional bias of the LLM judge. \textit{Wins}, \textit{ties}, and \textit{losses} are scored as $1$, $0.5$, and $0$, respectively. The score of $\tau_i$ under a rubric is the mean over its incident edges, and averaging across the active rubrics yields the composite process score $s_i^{\text{rubric}}$.

\paragraph{Centering and reward integration.}
Before entering the reward, ARBOR applies a per-batch variance filter. 
If an active rubric's within-group variance falls below $\delta_v$ on a batch, its scores are discarded to avoid injecting uninformative noise.
The scores from the filtered active rubrics are then centered within each query-group,
\begin{equation}
\label{eq:centering}
\tilde{s}_i^{\text{rubric}}
=
s_i^{\text{rubric}}
-
\frac{1}{|\mathcal{G}_q|}
\sum_{j \in \mathcal{G}_q} s_j^{\text{rubric}},
\end{equation}
and the final reward uses additive shaping,
\begin{equation}
\label{eq:reward}
\begin{aligned}
R_i^{\text{total}} &= R_i^{\text{base}} + \lambda \cdot \tilde{s}_i^{\text{rubric}}, \\
R_i^{\text{base}} &=
\begin{cases}
R_i^{\text{F1}} & \text{if format valid} \\
-1 & \text{otherwise}
\end{cases},
\end{aligned}
\end{equation}
where $R_i^{\text{F1}}$ is the token-level F1 of the final answer and $\lambda$ is fixed throughout training. 
Group centering gives the rubric term zero mean within each query-group, and negative centered scores are further attenuated by a factor $\alpha$, encouraging stronger search
behavior without over-penalizing merely less-preferred trajectories.
For format-invalid trajectories, the rubric term is forced to zero, preserving the format penalty imposed by $R_i^{\text{base}}$.

\section{Experiments}
\label{sec:experiments}

\begin{table*}[t]
\centering
\resizebox{\textwidth}{!}{%
\begin{tabular}{l ccc ccc ccc ccc ccc}
\toprule
\multirow{2}{*}{\textbf{Method}} & \multicolumn{3}{c}{\textbf{Bamboogle}} & \multicolumn{3}{c}{\textbf{HotpotQA}} & \multicolumn{3}{c}{\textbf{MuSiQue}} & \multicolumn{3}{c}{\textbf{2Wiki}} & \multicolumn{3}{c}{\textbf{Avg.}} \\
\cmidrule(lr){2-4} \cmidrule(lr){5-7} \cmidrule(lr){8-10} \cmidrule(lr){11-13} \cmidrule(lr){14-16}
 & EM & F1 & Acc. & EM & F1 & Acc. & EM & F1 & Acc. & EM & F1 & Acc. & EM & F1 & Acc. \\
\midrule
\textbf{Qwen3-4B} & & & & & & & & & & & & & & & \\
\quad + TIR Prompting & 46.4 & 55.3 & 61.1 & 30.4 & 38.5 & 46.2 & 9.2 & 16.2 & 19.8 & 32.8 & 42.3 & 47.2 & 29.7 & 38.1 & 43.6 \\
\quad + GRPO & 61.6 & 69.8 & 71.0 & \underline{44.9} & \underline{54.5} & \underline{57.4} & \textbf{17.0} & \textbf{26.9} & \underline{26.5} & \underline{60.9} & \underline{68.3} & 67.6 & \underline{46.1} & 54.9 & 55.6 \\
\quad + DAPO & \underline{63.7} & \underline{72.7} & \underline{74.2} & 44.2 & 53.9 & 57.0 & 16.4 & \underline{25.9} & 25.2 & 60.2 & 67.7 & \underline{68.1} & \underline{46.1} & \underline{55.1} & \underline{56.1} \\
\rowcolor{blue!8}
\quad + ARBOR & \textbf{65.1} & \textbf{73.0} & \textbf{76.2} & \textbf{45.4} & \textbf{55.6} & \textbf{59.7} & \textbf{17.0} & 25.7 & \textbf{27.8} & \textbf{64.2} & \textbf{72.4} & \textbf{74.6} & \textbf{47.9} & \textbf{56.7} & \textbf{59.6} \\
\midrule
\textbf{Qwen3-8B} & & & & & & & & & & & & & & & \\
\quad + TIR Prompting & 31.2 & 38.1 & 41.3 & 15.7 & 20.2 & 22.9 & 6.2 & 9.8 & 11.6 & 9.7 & 13.8 & 15.1 & 15.7 & 20.5 & 22.7 \\
\quad + GRPO & 61.0 & 69.4 & 72.2 & 44.4 & \underline{55.0} & \underline{57.8} & \underline{17.4} & \underline{26.7} & \underline{26.6} & \textbf{64.8} & \underline{71.0} & \underline{71.1} & 46.9 & \underline{55.5} & \underline{56.9} \\
\quad + DAPO & \underline{61.9} & \underline{70.1} & \underline{73.1} & \underline{45.8} & 54.9 & 57.7 & 17.0 & 25.8 & 26.2 & 63.8 & 69.7 & 69.6 & \underline{47.1} & 55.1 & 56.7 \\
\rowcolor{blue!8}
\quad + ARBOR & \textbf{65.3} & \textbf{72.9} & \textbf{76.5} & \textbf{47.8} & \textbf{58.6} & \textbf{62.9} & \textbf{18.3} & \textbf{26.8} & \textbf{29.5} & \underline{64.2} & \textbf{73.3} & \textbf{75.3} & \textbf{48.9} & \textbf{57.9} & \textbf{61.1} \\
\midrule
\textbf{Qwen3-14B} & & & & & & & & & & & & & & & \\
\quad + TIR Prompting & 42.1 & 50.3 & 49.9 & 27.8 & 34.6 & 39.2 & 8.6 & 16.0 & 17.4 & 26.1 & 31.9 & 34.4 & 26.2 & 33.2 & 35.2 \\
\quad + GRPO & 61.9 & 71.5 & 72.8 & \underline{48.1} & \underline{58.4} & \underline{62.3} & \textbf{19.4} & \textbf{29.9} & \textbf{31.6} & \textbf{69.8} & \underline{76.2} & \underline{76.4} & \underline{49.8} & \underline{59.0} & \underline{60.8} \\
\quad + DAPO & \underline{65.8} & \underline{74.1} & \underline{74.9} & 46.2 & 57.5 & 61.1 & 17.2 & 27.1 & 27.3 & 67.6 & 73.4 & 72.9 & 49.2 & 58.0 & 59.1 \\
\rowcolor{blue!8}
\quad + ARBOR & \textbf{67.5} & \textbf{75.8} & \textbf{78.6} & \textbf{48.6} & \textbf{59.1} & \textbf{63.6} & \underline{19.1} & \underline{28.7} & \textbf{31.6} & \textbf{69.8} & \textbf{76.3} & \textbf{77.3} & \textbf{51.3} & \textbf{60.0} & \textbf{62.8} \\
\bottomrule
\end{tabular}%
}
\caption{\textbf{Overall performance on 4 knowledge-intensive reasoning tasks.} ARBOR achieves the best average EM, F1 and LLM-judge accuracy at all three scales. The top two outcomes in each column are \textbf{bolded} and \underline{underlined}.}
\label{tab:main-results}
\end{table*}

\subsection{Experimental Setup}
\label{sec:setup}

\paragraph{Datasets.}
Our training follows a two-stage SFT-then-RL procedure. SFT uses a search-tool subset of Tool-Star~\citep{dong2025toolstar} and STILL~\citep{rucaibox2025still}, filtered to remove Python-tool examples, yielding approximately 16K training examples. RL training uses 2K QA examples randomly sampled from the ARPO Deep Reasoning Tasks~\citep{dong2025arpo}. Evaluation is on four multi-hop QA benchmarks: Bamboogle~\citep{press-etal-2023-measuring}, HotpotQA~\citep{yang-etal-2018-hotpotqa}, MuSiQue~\citep{trivedi-etal-2022-musique} and 2WikiMultiHopQA~\citep{ho-etal-2020-constructing}. The test split strategy follows the convention of ARPO~\citep{dong2025arpo}. Benchmark details are provided in Appendix~\ref{sec:app-benchmarks}.

\paragraph{Baselines.}
We use Qwen3-8B~\citep{yang2025qwen3} as the main backbone, with Qwen3-4B and Qwen3-14B for cross-scale verification, comparing three methods on each backbone: (1) TIR Prompting, which injects the search-tool specification but performs no RL fine-tuning; (2) GRPO~\citep{shao2024deepseekmath}, which uses base reward as the sole reward; and 
(3) DAPO~\citep{yu2025dapo}, which discards outcome-homogeneous query-groups from policy updates, providing a contrast on how to handle the zero-gradient groups that ARBOR exploits.

\paragraph{Implementation Details.}
All training runs on a single node with 8$\times$H100 GPUs using Slime~\citep{thudm2026slime} as the RL framework. The search tool is Google Search, returning 10 results per query. The base reward is token-level F1 with a format penalty. For ARBOR, induction, consolidation, and pairwise judging all call Qwen3-Plus\footnote{Model id: qwen-plus-2025-04-28. Accessed: 2026-05}~\citep{alibabacloud2026qwenplus} as the external LLM. Consolidation deduplication uses BGE-M3~\citep{chen2024bgem3} sentence embeddings, and the reward shaping coefficient is fixed to $\lambda = 0.1$. Remaining hyperparameters are listed in Appendix~\ref{sec:app-hparams}.

\paragraph{Evaluation Metrics.}
We report Exact Match (EM), token-level F1, and LLM-judge accuracy with Qwen3-Plus as the judge. All numbers are pass@1 averaged over 5 independent evaluation runs to reduce sampling variance. The LLM-judge prompt is given in Appendix~\ref{sec:app-prompt-eval}.

\subsection{Main Results}

Table~\ref{tab:main-results} presents the main results across four multi-hop QA benchmarks. ARBOR achieves the best average performance at every model scale. We highlight three observations from the results.

\paragraph{ARBOR consistently improves over outcome-only baselines.}
ARBOR improves average LLM-judge accuracy over GRPO by \textbf{4.0}, \textbf{4.2}, and \textbf{2.0} points at 4B, 8B, and 14B, and also uniformly outperforms DAPO at all three scales. The improvements are consistent across evaluation metrics, with gains observed not only in LLM-judge accuracy but also in EM and F1. 
TIR Prompting performs substantially worse than all RL-trained variants, confirming that tool access alone is insufficient without policy optimization. Its non-monotonic scaling is mainly caused by answer-format failures under the strict evaluator, as diagnosed in Appendix~\ref{sec:app-tir-format}.

\paragraph{Exploiting outcome-homogeneous groups outperforms discarding them.}
DAPO's gains over GRPO are inconsistent across scales, turning negative at 14B
($-$1.7 points), while ARBOR consistently outperforms GRPO at
all scales. An important difference is how the methods handle outcome-homogeneous
groups: DAPO discards them entirely from policy updates, whereas ARBOR retains
them and applies reusable process rubrics to provide within-group discrimination.
ARBOR's lead over DAPO is uniformly large (\textbf{+3.5} to \textbf{+4.4} points across all
backbones), confirming that exploiting these groups through process reward is
more effective than discarding them. 

\paragraph{Process supervision translates into semantic answer gains.}
ARBOR's largest gains appear on LLM-judge accuracy: +4.0, +4.2, and +2.0 points over GRPO at 4B, 8B, and 14B, exceeding the corresponding EM gains (+1.8, +2.0, +1.5) and F1 gains (+1.8, +2.4, +1.0).  
This pattern is consistent with ARBOR's design: rubrics reward better query formulation, evidence use, and answer synthesis rather than final-answer string overlap. 
The result suggests that reusable process supervision improves answer quality in ways that are better captured by semantic evaluation than by lexical-overlap metrics alone.
To ensure these gains are not evaluator-specific, Appendix~\ref{sec:app-cross-judge} verifies the same trend under DeepSeek-V4-Pro~\citep{deepseek2026v4pro} as an independent evaluator.

\subsection{No-Gradient Group Rescue}
\label{sec:no-gradient-rescue}

\begin{table}[t]
    \centering
    \scriptsize
    \setlength{\tabcolsep}{4pt}
    \resizebox{\columnwidth}{!}{%
    \begin{tabular}{llcccc}
        \toprule
        \textbf{Model} & \textbf{Reward} & \textbf{All-corr.}$\downarrow$ & \textbf{All-wrong}$\downarrow$ & \textbf{Mixed-uni.}$\downarrow$ & \textbf{Total}$\downarrow$ \\
        \midrule
        \multirow{3}{*}{Qwen3-4B}
            & $R^{\text{base}}$  & 26.9\% & 14.7\% & 2.9\% & 44.5\% \\
            & $R^{\text{total}}$ & 22.2\% & 6.0\% & 2.2\% & 30.4\% \\
            & \cellcolor{blue!8}$\Delta_{\text{rel}}$ & \cellcolor{blue!8}$-17.5\%$ & \cellcolor{blue!8}$-59.2\%$ & \cellcolor{blue!8}$-24.1\%$ & \cellcolor{blue!8}$-31.7\%$ \\
        \midrule
        \multirow{3}{*}{Qwen3-8B}
            & $R^{\text{base}}$  & 23.5\% & 14.3\% & 2.9\% & 40.7\% \\
            & $R^{\text{total}}$ & 16.2\% & 5.6\% & 1.6\% & 23.4\% \\
            & \cellcolor{blue!8}$\Delta_{\text{rel}}$ & \cellcolor{blue!8}$-31.1\%$ & \cellcolor{blue!8}$-60.8\%$ & \cellcolor{blue!8}$-44.8\%$ & \cellcolor{blue!8}$-42.5\%$ \\
        \midrule
        \multirow{3}{*}{Qwen3-14B}
            & $R^{\text{base}}$  & 30.1\% & 14.3\% & 3.5\% & 48.0\% \\
            & $R^{\text{total}}$ & 23.9\% & 6.6\% & 2.0\% & 32.5\% \\
            & \cellcolor{blue!8}$\Delta_{\text{rel}}$ & \cellcolor{blue!8}$-20.6\%$ & \cellcolor{blue!8}$-53.8\%$ & \cellcolor{blue!8}$-42.9\%$ & \cellcolor{blue!8}$-32.3\%$ \\
        \bottomrule
    \end{tabular}%
    }
    \caption{\textbf{Reward-homogeneous group fractions during ARBOR training.} Groups are measured on $R^{\text{base}}$ (before rubric shaping) and $R^{\text{total}}$ (after rubric shaping).}
    \label{tab:no-gradient}
\end{table}

A direct test of the motivating claim from Section~\ref{sec:introduction} is whether rubric scoring provides within-group discrimination on groups where outcome-only reward produces zero gradient. Table~\ref{tab:no-gradient} reports the fraction of groups with zero within-group reward variance, measured first under $R^{\text{base}}$ alone and then under $R^{\text{total}}$. A drop indicates that rubric scoring introduced reward discrimination that the outcome signal could not provide. Groups are further split by outcome pattern: \emph{all-correct} groups have F1=1 on every trajectory with valid format, \emph{all-wrong} groups have F1=0 on every trajectory including format-invalid cases, and \emph{mixed-uniform} groups share the same partial F1 across all trajectories.

Rubric scoring reduces all three types at every scale. The effect concentrates on \emph{all-wrong} groups, where the relative reduction reaches \textbf{54--61\%}. These groups represent queries the policy has not yet learned to solve, and rubric scoring directly helps here by distinguishing more promising search strategies from less promising ones, providing within-group discrimination that the base reward cannot. The reduction on \emph{all-correct} groups is smaller (\textbf{18--31\%}), as the active rubrics are primarily induced from success-failure contrasts and less attuned to the subtler process differences among trajectories that all succeed. Overall, rubric scoring converts \textbf{32--42\%} of outcome-homogeneous groups into ones with nonzero within-group reward variance, enabling policy learning on these groups.

\subsection{Effect of Reusable Rubric Memory}
\label{sec:rar-memory}

\begin{figure}[t]
    \centering
    \includegraphics[width=\linewidth]{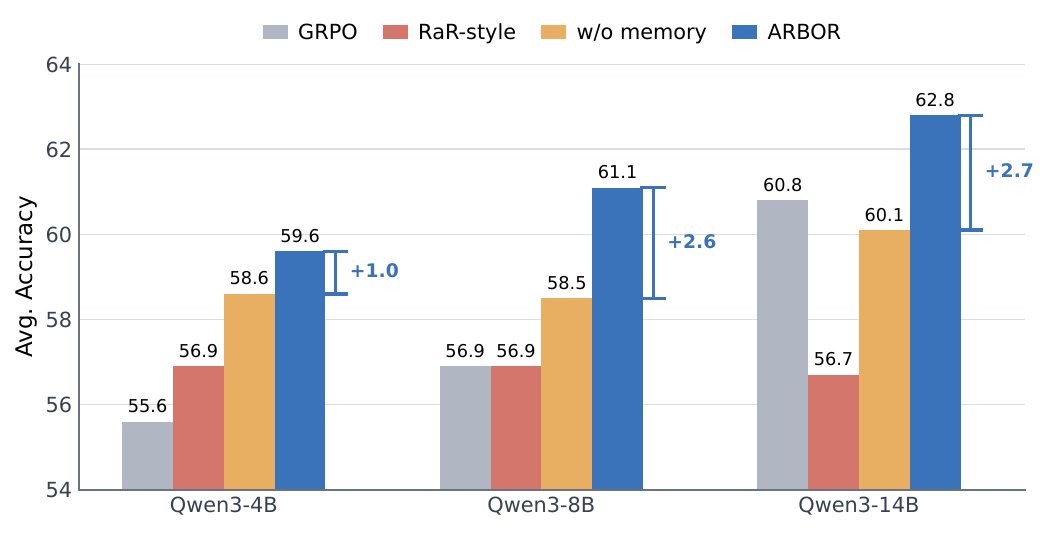}
    \caption{\textbf{Effect of reusable rubric memory.} Average LLM-judge accuracy across four benchmarks.}
    \label{fig:rar-single-use}
\end{figure}

To isolate the contribution of reusable rubric memory, Figure~\ref{fig:rar-single-use} compares ARBOR with two no-memory rubric variants. RaR-style, adapted from Rubrics-as-Rewards~\citep{gunjal2025rar}, uses static query-specific rubrics generated from the question and reference answer to score rollouts of the corresponding query. The w/o memory variant uses ARBOR's contrastive induction but discards each rubric after scoring its source query-group. 
Detailed experimental setup and full results are provided in Appendix~\ref{sec:app-full-ablation}.

No-memory rubric feedback is not sufficient to explain ARBOR's gains. RaR-style is weak and unstable: it slightly improves average accuracy over GRPO at 4B (+1.3 points), matches GRPO at 8B, and falls substantially below GRPO at 14B (-4.1 points). The w/o memory variant is stronger than RaR-style at all three scales, showing that contrastive process rubrics are more useful than reference-conditioned instance checklists. However, it still lacks the stability of reusable memory. ARBOR outperforms w/o memory by \textbf{1.0}, \textbf{2.6}, and \textbf{2.7} accuracy points at 4B, 8B, and 14B, and outperforms RaR-style by \textbf{2.7}, \textbf{4.2}, and \textbf{6.1} points. These margins indicate that the main benefit is not merely adding rubric-shaped reward, but converting local observations into common process criteria that can be filtered and reused across later query-groups.

\subsection{Common-Rubric Reuse}
\label{sec:reuse-diagnostics}

\begin{table}[t]
    \centering
    \footnotesize
    \setlength{\tabcolsep}{5pt}
    \renewcommand{\arraystretch}{1.15}
    \begin{tabularx}{\columnwidth}{>{\raggedright\arraybackslash}Xc}
        \toprule
        \textbf{Common rubric} & \textbf{Uses} \\
        \midrule
        Precise Entity-Attribute Targeting with Canonical Framing & 101 \\
        Evidence-Sufficiency-Guided Termination with Cross-Validated Convergence & 84 \\
        Precision in Relational Query Formulation & 55 \\
        Constraint-Integrated Entity-Attribute Binding & 54 \\
        Constraint-Consistent Interpretation of Question Semantics and Intent & 54 \\
        \bottomrule
    \end{tabularx}
    \caption{\textbf{Top-5 most reused common rubrics during training.} Uses denotes the number of query-groups each rubric is activated to score.}
    \label{tab:reused-rubrics}
\end{table}

A central claim of ARBOR is that consolidated common rubrics serve as reusable cross-query process standards rather than one-shot rewards. We verify this by ranking all common rubrics produced during training by the number of query-groups each one is activated to score. The top 5\% of rubrics by this ranking account for 20\% of all scoring events and the top 20\% account for 44\%, as detailed in Appendix~\ref{sec:app-reuse-lorenz}. This confirms that a small core of high-quality rubrics emerges and consistently drives process supervision across diverse queries.

Table~\ref{tab:reused-rubrics} lists the most reused rubrics, all of which encode generic process behaviors such as precise entity-attribute targeting and evidence-guided termination. None references an entity or fact specific to any individual query, confirming that the consolidation step successfully abstracts query-local drafts into genuinely cross-query standards. A concrete consolidation case is shown in Appendix~\ref{sec:app-cases}.

\subsection{Hyperparameter Sensitivity}

\begin{figure}[t]
    \centering
    \includegraphics[width=\columnwidth]{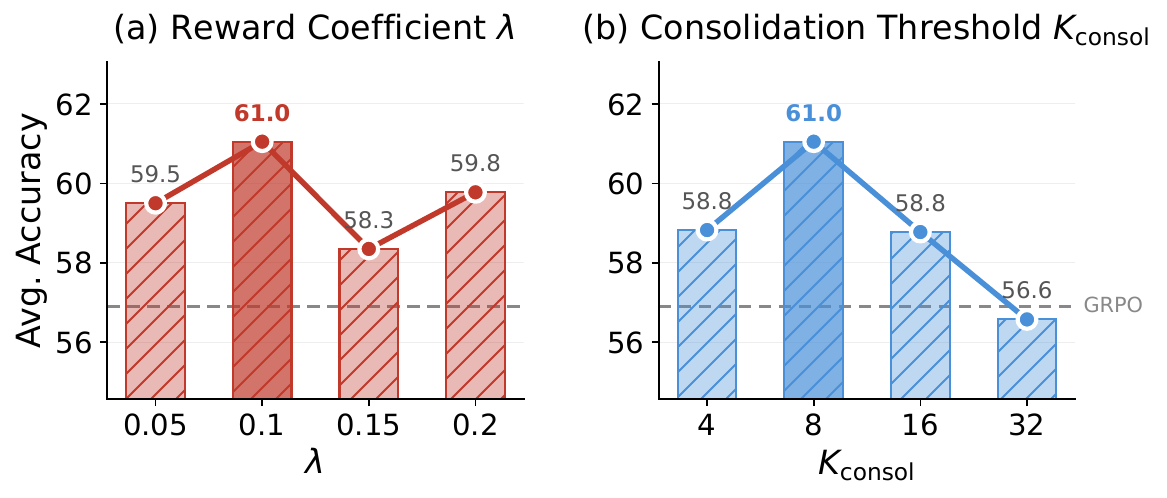}
    \caption{\textbf{Hyperparameter sensitivity on $\lambda$ and $K_{\text{consol}}$.} The dashed line marks the GRPO baseline.}
    \label{fig:hparam-sensitivity}
\end{figure}

ARBOR adds a process reward term and a rubric memory mechanism, so we evaluate sensitivity to the two hyperparameters most tied to them: the reward coefficient $\lambda$, which controls the strength of the process signal, and the consolidation threshold $K_{\text{consol}}$, which controls how quickly local drafts are converted into reusable common rubrics.

Figure~\ref{fig:hparam-sensitivity} shows the results. Panel~(a) varies $\lambda$. Performance peaks at $\lambda=0.1$; smaller values underweight the rubric signal, while larger values allow noisy rubric scores to interfere with the outcome gradient. All four settings outperform GRPO, indicating that ARBOR is robust over a reasonable range of $\lambda$. Panel~(b) varies $K_{\text{consol}}$. This parameter trades off reliability against timeliness: small values trigger consolidation from limited evidence, while large values increase the lag between induction and activation so that rubrics reflect an outdated policy distribution. All values in $\{4, 8, 16\}$ outperform the baseline, with the best at $K_{\text{consol}}=8$. At $K_{\text{consol}}=32$, the consolidation lag causes rubrics to fall behind the evolving policy and performance drops.

\section{Conclusion}

In this work, we introduced ARBOR, a reusable process-reward framework designed to address the limitations of outcome-only rewards in search-agent RL. ARBOR induces natural-language process rubrics from contrastive trajectories, consolidates them into a reusable cross-query memory, and manages the memory through an admission--consolidation--retirement lifecycle. This design jointly achieves process-level supervision, cross-query consistency, and co-evolution with the policy. Experiments across three model scales and four multi-hop QA benchmarks show that ARBOR consistently improves over outcome-only RL baselines, while ablations and reuse analyses confirm the importance of reusable rubric memory. These results suggest that process knowledge in agent training need not be ephemeral: when accumulated into reusable criteria that co-evolve with the policy, it provides a persistent and growing source of supervision that outcome signals alone cannot offer.

\section*{Limitations}

First, ARBOR relies on an external LLM for rubric induction, consolidation, and pairwise judging, so the quality of rubric supervision is tied to the capability of the external model. Models with stronger reasoning capacity may yield more precise rubrics and finer process discrimination. Second, our experiments validate ARBOR on multi-hop QA with a single search tool. Since the rubric memory is task-agnostic by construction, extending it to broader tool-use settings such as code generation and mathematical reasoning with retrieval is a straightforward direction.

\bibliography{custom}

\appendix

\section{Method Details}
\label{sec:app-method-details}

\subsection{Contrastive Induction Corner Cases}
\label{sec:app-induction}

This section records the rules ARBOR follows when correctness scores within a query-group tie and when the two contrast pairs collapse into one, complementing Section~\ref{sec:induction}.

\paragraph{Tie-breaking for anchor selection.}
When multiple trajectories share the highest F1, $\tau^+$ is the one with the shortest response, and the same secondary key applies when selecting $\tau^-_{\text{hard}}$ and $\tau^-_{\text{worst}}$.

\paragraph{Collapsed contrast pairs.}
When only one trajectory falls strictly below $\tau^+$ in F1, $\tau^-_{\text{hard}}$ and $\tau^-_{\text{worst}}$ resolve to the same trajectory; in this case only the $(\tau^+, \tau^-_{\text{hard}})$ pair is sent to the LLM.
When no trajectory falls below $\tau^+$, the group has no correctness divergence and induction is skipped entirely.

\paragraph{Homogeneous groups.}
For all-correct or all-wrong groups, new rubric generation is skipped entirely; existing common rubrics in $\mathcal{P}$ still score the group for reward shaping.
For groups in which all trajectories share the same intermediate F1 value, no contrast pairs can be formed, and induction falls back to sending all trajectories to the LLM without positive/negative labels, letting it identify process differences directly.
Induction is also skipped when fewer than two format-valid trajectories exist in the group.

\section{Training and Evaluation Details}
\label{sec:app-setup}

\subsection{Evaluation Benchmarks}
\label{sec:app-benchmarks}

We evaluate on the following four multi-hop QA benchmarks:
\begin{itemize}
    \item \textbf{HotpotQA}~\citep{yang-etal-2018-hotpotqa} is a Wikipedia-based multi-hop QA benchmark that requires aggregating evidence across multiple supporting documents to answer a question. It is characterized by two representative question types: reasoning over sentence-level supporting facts across documents, and comparison questions that contrast properties of two entities.
    \item \textbf{2WikiMultiHopQA}~\citep{ho-etal-2020-constructing} is a multi-hop QA dataset constructed from both Wikipedia and Wikidata, combining unstructured text with structured knowledge to support multi-step reasoning. It provides explicit evidence and reasoning paths, ensuring that answering each question requires genuine multi-hop inference.
    \item \textbf{MuSiQue}~\citep{trivedi-etal-2022-musique} is designed to resist shortcut solutions by composing connected single-hop questions into multi-hop ones, such that proper multi-hop reasoning is required by construction. Unlike earlier benchmarks that can often be resolved with partial evidence, MuSiQue ensures that each step of the reasoning chain is necessary to reach the correct answer.
    \item \textbf{Bamboogle}~\citep{press-etal-2023-measuring} is a hand-crafted benchmark of 125 two-hop compositional questions designed to be difficult for standard search engines to answer directly, while the supporting evidence for each question can be found in Wikipedia. It evaluates models on varied compositional questions that require combining information across multiple retrieval steps.
\end{itemize}

Following Tool-Star~\citep{dong2025toolstar} and ARPO~\citep{dong2025arpo}, we evaluate on fixed held-out test splits: 200 examples for HotpotQA, 200 for 2WikiMultiHopQA, 200 for MuSiQue, and 125 for Bamboogle.

\subsection{SFT and RL Training}
\label{sec:app-training}

\paragraph{Supervised Fine-Tuning.}
Starting from Tool-Star~\citep{dong2025toolstar} and STILL~\citep{rucaibox2025still},
we filter out all examples that invoke Python as a tool and retain only
search-tool interactions, yielding approximately 16K training examples.
Fine-tuning is conducted with LLaMAFactory~\citep{zheng-etal-2024-llamafactory}
using full-parameter optimization, DeepSpeed ZeRO Stage~3~\citep{rasley2020zero},
and FlashAttention-2~\citep{dao2024flashattention2}.
We train for 3 epochs with a maximum sequence length of 15{,}000 tokens,
an effective batch size of 16, a learning rate of $7\times10^{-6}$ with
cosine decay and 10\% linear warmup, in BF16 precision.

\paragraph{Reinforcement Learning.}
RL training runs for 60 rollout steps using Slime~\citep{thudm2026slime}.
We sample 8 trajectories per prompt with a rollout batch size of 32,
sampling temperature 0.7, a maximum context length of 36{,}864 tokens,
and a maximum response length of 8{,}192 tokens.
Each trajectory is limited to at most 12 agent turns and 10 search calls.
The policy is updated with GRPO with token-level importance sampling (TIS),
KL and entropy coefficients set to 0, and symmetric clipping with
$\epsilon_{\text{clip}}=0.2$.
Optimization uses Adam with learning rate $1\times10^{-6}$ and weight decay 0.01.
\subsection{ARBOR Hyperparameters}
\label{sec:app-hparams}

Table~\ref{tab:app-arbor-hparams} lists the default ARBOR hyperparameters
used in the main experiments.
We elaborate on the key design choices below.

\begin{table}[t]
    \centering
    \small
    \setlength{\tabcolsep}{6pt}
    \renewcommand{\arraystretch}{1.05}
    \begin{tabular}{p{0.56\columnwidth}cc}
        \toprule
        Hyperparameter & Notation & Default \\
        \midrule
        Rubric shaping coefficient         & $\lambda$                & 0.1  \\
        Consolidation trigger              & $K_{\text{consol}}$      & 8    \\
        Admission variance threshold       & $\delta_v$               & 0.05 \\
        Correlation threshold              & $\rho_{\min}$            & 0.0  \\
        Retirement tolerance               & $\kappa$                 & 5    \\
        Negative score attenuation factor  & $\alpha$                 & 0.25 \\
        Maximum common pool size           & $|\mathcal{P}|_{\max}$   & 6    \\
        Active common rubrics per step     & ---                      & 2    \\
        Deduplication similarity threshold & ---                      & 0.9  \\
        \bottomrule
    \end{tabular}
    \caption{Default ARBOR hyperparameters used in the main experiments.}
    \label{tab:app-arbor-hparams}
\end{table}

$K_{\text{consol}}=8$ is the number of candidate drafts that must accumulate in $\mathcal{D}$ before a consolidation round is triggered.
$\kappa=5$ is the maximum number of consecutive low-variance activations a common rubric is allowed before retirement.
The deduplication similarity threshold is set to 0.9, above which a newly produced common rubric is considered a duplicate of an existing one under BGE-M3~\citep{chen2024bgem3} cosine similarity.

\section{Additional Experimental Results}
\label{sec:app-additional-results}

\subsection{Format Sensitivity of TIR Prompting}
\label{sec:app-tir-format}

The official evaluation used in Table~\ref{tab:main-results} is strict-format gated: a response is scored only if it produces a non-empty prediction and terminates with the required final-answer format.
If the termination reason indicates a formatting or finalization failure, all official metrics are set to zero, including EM, F1, and LLM-judge accuracy.
This protocol matches the training reward, where finalization is part of the task rather than a post-hoc parsing detail.

To diagnose the non-monotonic TIR Prompting results in the main table, Table~\ref{tab:app-tir-format} separates formatting validity from answer quality.
\emph{Official} accuracy follows the strict protocol above, \emph{valid-only} accuracy evaluates only responses that satisfy the required format, and \emph{recovery} is the gap between the two LLM accuracies.

\begin{table*}[t]
    \centering
    \small
    \setlength{\tabcolsep}{12pt}
    \renewcommand{\arraystretch}{1.15}
    \begin{tabular}{lcccc}
        \toprule
        \textbf{Model} & \textbf{Valid Rate} & \textbf{Official LLM Acc.} & \textbf{Valid-only LLM Acc.} & \textbf{Recovery} \\
        \midrule
        Qwen3-4B  & 89.7 & 43.6 & 48.4 & \textbf{+4.8} \\
        Qwen3-8B  & 45.1 & 22.7 & 48.6 & \textbf{+25.9} \\
        Qwen3-14B & 68.5 & 35.2 & 52.4 & \textbf{+17.2} \\
        \bottomrule
    \end{tabular}
    \caption{\textbf{Format sensitivity of TIR Prompting.} All values are percentages averaged over four benchmarks and five evaluation runs.}
    \label{tab:app-tir-format}
\end{table*}

Table~\ref{tab:app-tir-format} shows that the non-monotonic TIR results are mainly driven by formatting failures.
Qwen3-8B and Qwen3-14B have much lower valid rates than Qwen3-4B, which sharply lowers their official LLM-judge accuracies.
On valid responses, however, they reach 48.6\% and 52.4\% LLM-judge accuracy, comparable to or higher than Qwen3-4B.
Thus, the apparent weakness of larger TIR backbones primarily reflects unreliable prompt-only finalization under a strict protocol, not weaker search reasoning.
We keep the strict official scores in the main table because correct finalization is part of the tool-use task and is also enforced during RL training.

\subsection{Cross-Judge Robustness}
\label{sec:app-cross-judge}

To check whether the main LLM-judge trends depend on the same model family used by ARBOR's reward construction, we re-evaluate all final answers with DeepSeek-V4-Pro~\citep{deepseek2026v4pro} as an independent post-hoc judge.
DeepSeek-V4-Pro is not used for rubric induction, consolidation, pairwise scoring, or reward construction.
Table~\ref{tab:app-cross-judge} compares average LLM-judge accuracy under Qwen3-Plus and DeepSeek-V4-Pro.
The independent judge preserves the same ranking: ARBOR remains the best method at all three scales, and its gains over GRPO remain close to the Qwen3-Plus gains.

\begin{table*}[t]
\centering
\small
\setlength{\tabcolsep}{6pt}
\begin{tabular}{llccc|cc}
\toprule
\multirow{2}{*}{\textbf{Backbone}} & \multirow{2}{*}{\textbf{Method}}
& \multicolumn{3}{c|}{\textbf{Judge Accuracy}}
& \multicolumn{2}{c}{\textbf{Gain over GRPO}} \\
\cmidrule(lr){3-5}\cmidrule(lr){6-7}
& & \textbf{Qwen3-Plus} & \textbf{DeepSeek} & \textbf{$\Delta$}
& \textbf{Qwen3-Plus} & \textbf{DeepSeek} \\
\midrule
\multirow{4}{*}{Qwen3-4B}
& TIR Prompting & 43.6 & 42.4 & $-1.2$ & $-12.0$ & $-12.7$ \\
& GRPO          & 55.6 & 55.1 & $-0.5$ & 0.0     & 0.0 \\
& DAPO          & 56.1 & 55.6 & $-0.5$ & $+0.5$  & $+0.5$ \\
& \cellcolor{blue!8}ARBOR         & \cellcolor{blue!8}\textbf{59.6} & \cellcolor{blue!8}\textbf{58.7} & \cellcolor{blue!8}$-0.9$ & \cellcolor{blue!8}\textbf{$+4.0$} & \cellcolor{blue!8}\textbf{$+3.6$} \\
\midrule
\multirow{4}{*}{Qwen3-8B}
& TIR Prompting & 22.7 & 22.0 & $-0.7$ & $-34.2$ & $-34.6$ \\
& GRPO          & 56.9 & 56.5 & $-0.4$ & 0.0     & 0.0 \\
& DAPO          & 56.7 & 55.7 & $-1.0$ & $-0.2$  & $-0.8$ \\
& \cellcolor{blue!8}ARBOR         & \cellcolor{blue!8}\textbf{61.1} & \cellcolor{blue!8}\textbf{60.6} & \cellcolor{blue!8}$-0.5$ & \cellcolor{blue!8}\textbf{$+4.2$} & \cellcolor{blue!8}\textbf{$+4.0$} \\
\midrule
\multirow{4}{*}{Qwen3-14B}
& TIR Prompting & 35.2 & 34.6 & $-0.6$ & $-25.6$ & $-25.3$ \\
& GRPO          & 60.8 & 59.9 & $-0.9$ & 0.0     & 0.0 \\
& DAPO          & 59.1 & 58.6 & $-0.5$ & $-1.7$  & $-1.3$ \\
& \cellcolor{blue!8}ARBOR         & \cellcolor{blue!8}\textbf{62.8} & \cellcolor{blue!8}\textbf{61.8} & \cellcolor{blue!8}$-1.0$ & \cellcolor{blue!8}\textbf{$+2.0$} & \cellcolor{blue!8}\textbf{$+1.9$} \\
\bottomrule
\end{tabular}
\caption{\textbf{Cross-judge robustness of average LLM-judge accuracy.}
Qwen3-Plus is the main evaluation judge, while DeepSeek denotes DeepSeek-V4-Pro used only as an independent post-hoc judge.
$\Delta$ denotes DeepSeek accuracy minus Qwen3-Plus accuracy. Gain over GRPO denotes the average accuracy difference from GRPO under the corresponding judge.
}
\label{tab:app-cross-judge}
\end{table*}

Overall, DeepSeek-V4-Pro is slightly stricter than Qwen3-Plus: its average scores are lower by 0.4--1.2 points across methods, but the drop is small and uniform.
Despite this stricter independent judge, the relative ordering is preserved and ARBOR's gains over GRPO remain nearly unchanged (+3.6, +4.0, and +1.9 points at 4B, 8B, and 14B).
Thus, the main LLM-judge conclusion is robust across judge families rather than driven by a Qwen3-Plus-specific preference.

\subsection{Setup and Full Results for Rubric Memory Ablation}
\label{sec:app-full-ablation}

Table~\ref{tab:app-full-ablation} provides the per-benchmark results for the rubric memory ablation discussed in Section~\ref{sec:rar-memory}. 

RaR-style is our search-QA adaptation of Rubrics-as-Rewards~\citep{gunjal2025rar}, instantiated as a single-use instance-specific rubric reward. For each training query, the external rubric-generation LLM generates 5--10 reference-conditioned rubric items from only the question and ground-truth answer, without seeing sampled rollout trajectories. During RL, the judge model scores each full agent trajectory under the query-specific rubric set and returns a holistic 1--10 rating; we normalize it to $[0,1]$, center it within the rollout group, and add it to the base F1/format reward with the same shaping coefficient as ARBOR. The generated rubrics are used only for their source query group and are never reused by later groups. 

The w/o memory variant instead keeps ARBOR's contrastive process-rubric induction from sampled trajectories, but discards each induced rubric after scoring its source query-group. Thus both variants have no reusable memory, while differing in whether their rubrics are reference-conditioned instance criteria or contrastively induced process criteria.

\begin{table*}[t]
\centering
\resizebox{\textwidth}{!}{%
\begin{tabular}{l ccc ccc ccc ccc ccc}
\toprule
\multirow{2}{*}{\textbf{Method}} & \multicolumn{3}{c}{\textbf{Bamboogle}} & \multicolumn{3}{c}{\textbf{HotpotQA}} & \multicolumn{3}{c}{\textbf{MuSiQue}} & \multicolumn{3}{c}{\textbf{2Wiki}} & \multicolumn{3}{c}{\textbf{Avg.}} \\
\cmidrule(lr){2-4} \cmidrule(lr){5-7} \cmidrule(lr){8-10} \cmidrule(lr){11-13} \cmidrule(lr){14-16}
 & EM & F1 & Acc. & EM & F1 & Acc. & EM & F1 & Acc. & EM & F1 & Acc. & EM & F1 & Acc. \\
\midrule
\textbf{Qwen3-4B} & & & & & & & & & & & & & & & \\
\quad + GRPO & 61.6 & 69.8 & 71.0 & 44.9 & 54.5 & 57.4 & 17.0 & 26.9 & 26.5 & 60.9 & 68.3 & 67.6 & 46.1 & 54.9 & 55.6 \\
\quad + RaR-style & 63.0 & 70.0 & 72.3 & 45.1 & 54.3 & 57.7 & 16.9 & 25.7 & 27.0 & 60.2 & 68.8 & 70.5 & 46.3 & 54.7 & 56.9 \\
\quad + w/o memory & 64.3 & 72.5 & 74.6 & 45.5 & 55.1 & 58.9 & 18.4 & 27.1 & 28.3 & 62.8 & 71.0 & 72.6 & 47.8 & 56.4 & 58.6 \\
\rowcolor{blue!8}
\quad + ARBOR & 65.1 & 73.0 & 76.2 & 45.4 & 55.6 & 59.7 & 17.0 & 25.7 & 27.8 & 64.2 & 72.4 & 74.6 & 47.9 & 56.7 & 59.6 \\
\midrule
\textbf{Qwen3-8B} & & & & & & & & & & & & & & & \\
\quad + GRPO & 61.0 & 69.4 & 72.2 & 44.4 & 55.0 & 57.8 & 17.4 & 26.7 & 26.6 & 64.8 & 71.0 & 71.1 & 46.9 & 55.5 & 56.9 \\
\quad + RaR-style & 60.0 & 67.8 & 70.7 & 43.9 & 54.5 & 57.4 & 17.3 & 25.5 & 28.1 & 65.1 & 71.3 & 71.3 & 46.6 & 54.8 & 56.9 \\
\quad + w/o memory & 64.2 & 72.0 & 74.2 & 47.0 & 57.2 & 60.9 & 16.9 & 25.7 & 28.0 & 62.4 & 69.7 & 70.8 & 47.6 & 56.2 & 58.5 \\
\rowcolor{blue!8}
\quad + ARBOR & 65.3 & 72.9 & 76.5 & 47.8 & 58.6 & 62.9 & 18.3 & 26.8 & 29.5 & 64.2 & 73.3 & 75.3 & 48.9 & 57.9 & 61.1 \\
\midrule
\textbf{Qwen3-14B} & & & & & & & & & & & & & & & \\
\quad + GRPO & 61.9 & 71.5 & 72.8 & 48.1 & 58.4 & 62.3 & 19.4 & 29.9 & 31.6 & 69.8 & 76.2 & 76.4 & 49.8 & 59.0 & 60.8 \\
\quad + RaR-style & 59.8 & 68.8 & 69.4 & 45.3 & 55.5 & 58.7 & 17.2 & 26.9 & 27.8 & 66.8 & 71.6 & 70.9 & 47.3 & 55.7 & 56.7 \\
\quad + w/o memory & 64.5 & 73.8 & 76.0 & 48.7 & 58.3 & 62.4 & 18.4 & 27.3 & 27.5 & 67.9 & 74.4 & 74.3 & 49.9 & 58.5 & 60.1 \\
\rowcolor{blue!8}
\quad + ARBOR & 67.5 & 75.8 & 78.6 & 48.6 & 59.1 & 63.6 & 19.1 & 28.7 & 31.6 & 69.8 & 76.3 & 77.3 & 51.3 & 60.0 & 62.8 \\
\bottomrule
\end{tabular}%
}
\caption{\textbf{Per-benchmark results for the rubric memory ablation.} Average columns report the mean over the four benchmarks; Section~\ref{sec:rar-memory} discusses the averaged trends.}
\label{tab:app-full-ablation}
\end{table*}

\subsection{Reuse Distribution}
\label{sec:app-reuse-lorenz}

\begin{figure}[t]
    \centering
    \includegraphics[width=\linewidth]{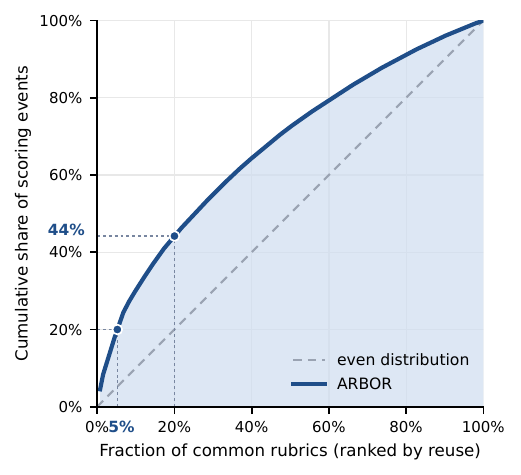}
    \caption{Cumulative share of scoring events covered by the top fraction of common rubrics, ranked by reuse count.}
    \label{fig:reuse-lorenz}
\end{figure}

Figure~\ref{fig:reuse-lorenz} plots the cumulative share of scoring events as a function of the fraction of common rubrics ranked by reuse. The curve sits well above the even-distribution diagonal. The top 5\% of rubrics cover 20\% of all scoring events, and the top 20\% cover 44\%. This concentration arises from the active selection mechanism in Section~\ref{sec:reward}, which preferentially activates rubrics with high cumulative correlation, so that a small core of validated rubrics accumulates most scoring events.

\section{Consolidation Case Study}
\label{sec:app-cases}

We illustrate the consolidation process with a concrete example. Table~\ref{tab:case-consolidation} shows three query-local candidate rubrics induced from different queries during RL training. Each candidate describes the same underlying process behavior, resolving the target entity with canonical naming before querying its attributes, but in different phrasings shaped by the specific query context.

The consolidation module receives these and other similar candidates as input, identifies the shared process pattern, and produces the common rubric \emph{Precise Entity-Attribute Targeting with Canonical Framing}. The resulting common rubric unifies three variant phrasings of the same process standard into a single reusable criterion, while the query-specific framing of each candidate is abstracted away. This abstraction is what enables the common rubric to be reused across 101 query-groups throughout training.

\begin{table*}[htbp]
    \centering
    \small
    \renewcommand{\arraystretch}{1.2}
    \begin{tabularx}{\textwidth}{>{\raggedright\arraybackslash}p{3.8cm}>{\raggedright\arraybackslash}X}
        \toprule
        \textbf{Source query} & \textbf{Candidate rubric (title and description)} \\
        \midrule
        Did Ashok Chavan and Peppino Tirri share the same nationality? &
        \textbf{Query Precision via Canonical Entity Anchoring.}
        A high-scoring response formulates its first search query using the canonical, widely attested name format of the target entity rather than vague role descriptors, enabling direct retrieval of authoritative biographic identifiers. \\
        \addlinespace
        What is the date of death of the performer of song Story Of A Life? &
        \textbf{Entity Disambiguation Before Attribute Search.}
        A high-scoring response first identifies and confirms the canonical entity using unambiguous, attribute-robust evidence before searching for biographical attributes like date of death---ensuring subsequent searches are grounded in the right person. \\
        \addlinespace
        The university which David R. Marchant was a member of in 1994 is located in what city? &
        \textbf{Precision-First Entity Identification Before Attribute Verification.}
        A high-scoring response treats the target as a concrete, disambiguated artifact requiring authoritative identification before querying attributes, prioritizing queries that resolve the entity's canonical identity in early steps. \\
        \midrule
        \multicolumn{2}{l}{\textbf{Consolidated common rubric:} \emph{Precise Entity-Attribute Targeting with Canonical Framing}} \\
        \bottomrule
    \end{tabularx}
    \caption{\textbf{Consolidation case study for reusable rubric memory.} Three query-local candidate rubrics from different queries describe the same entity-attribute targeting pattern in query-specific language, and the consolidation module abstracts them into a reusable common rubric.}
    \label{tab:case-consolidation}
\end{table*}

\section{Prompts}
\label{sec:app-prompts}

This section reproduces the four prompts used in ARBOR. Placeholders inside braces are filled at runtime, and ellipses denote repeated structures over the corresponding lists.

\subsection{Contrastive Induction}
\label{sec:app-prompt-induction}

The induction prompt has a fixed instruction header followed by a per-call payload that fills in the question itself, the optional ground-truth answer, the contrast pairs, and the existing rubrics; the input fields are listed at the end of the header. The full header is reproduced in Table~\ref{tab:prompt-induction}.

\begin{table*}[!t]
    \centering
    \fontsize{8pt}{10pt}\selectfont
    \renewcommand{\arraystretch}{0.9}
    \begin{tabular}{p{0.95\linewidth}}
    \toprule
        \rowcolor{gray!20}
        \textbf{Contrastive Induction Prompt} \\
    \midrule
        You are evaluating a \textbf{Deep Search Agent} --- a model that answers questions by iteratively searching the web, reading results, and synthesizing a final answer.\\
        Each response below contains the agent's full trajectory: its reasoning, search queries issued, search results received, and the final answer. \textbf{Answer correctness is already scored separately via strict-format-gated F1.}\\
        Format compliance is part of trajectory quality. A response with malformed tags, missing or unclosed $<$think$>$/$<$search$>$/$<$result$>$/$<$answer$>$ structure, multiple final answer blocks, missing boxed final answer, or truncation must not be treated as a high-quality positive example, even if a raw answer can be recovered and happens to match the ground truth.\\
        ~\\
        \textbf{Ideal Search Agent Vision.}
        An excellent Search Agent is defined by the quality of its reasoning and decisions at each step. The ideal agent:\\
        \quad - \textbf{Plans deliberately}: Understands the question deeply, identifies what information is needed, and decomposes complex questions into clear sub-goals before searching.\\
        \quad - \textbf{Searches precisely}: Formulates targeted queries using specific entity names and relationships to retrieve the right information directly.\\
        \quad - \textbf{Searches efficiently}: Avoids redundant queries, unnecessary re-verification of already-established facts, and circular search patterns that waste the search budget without adding new information once the key uncertainty has been resolved.\\
        \quad - \textbf{Adapts intelligently}: Searches further when genuine ambiguity, missing constraints, or conflicting evidence remain; synthesizes and answers when sufficient evidence has resolved the key uncertainty.\\
        \quad - \textbf{Answers confidently}: Produces a well-reasoned, correctly formatted final answer once the question can be resolved.\\
        \quad - \textbf{Finalizes cleanly}: Uses the required tool-call and answer format correctly, keeps tags properly paired and closed, and transitions cleanly from search to a single final answer without leaving unfinished structure.\\
        ~\\
        \textbf{Your Goal.}
        Identify \textbf{search strategy and reasoning process behaviors that CAUSE the agent to reach correct or incorrect answers}. The key question is: what did successful responses do differently in their \textit{process} (not their final answer) that led them to succeed? A good rubric targets a process behavior that, if improved, would lead to more correct answers. A bad rubric merely re-checks whether the answer content is correct (which F1 already does).\\
        \textbf{What to Look For.}
        Examine the provided contrast pairs and identify \textbf{specific process behaviors that caused some responses to succeed while others failed} (or that made some stronger responses clearly better-executed than weaker ones). Use the Ideal Search Agent Vision above as a reference for what good process looks like, but ground your rubrics in concrete observations from the given trajectories --- not generic best practices. Each rubric should capture a behavior dimension where the given responses \textbf{actually differ}, so that scoring produces meaningful variance across samples.\\
        ~\\
        \textbf{Rules.}\\
        \quad - \textbf{CRITICAL: Rubrics Must Target Process, Not Answer Content.} Do NOT generate rubrics that verify whether specific facts, names, dates, or relationships in the final answer are correct --- this is exactly what F1 already measures.\\
        \quad - \textbf{Format as Process Quality.} Treat clean finalization and valid interaction structure as part of process quality. Format-invalid or truncated responses may be used as low-quality anchors when they reveal unfinished reasoning, invalid tool-use structure, or failure to produce a single parseable final answer. However, avoid generating rubrics that only duplicate the strict format checker; prefer criteria that connect format discipline with reasoning, evidence use, search control, and answer finalization.\\
        \quad - \textbf{Using Ground Truth.} If a Ground Truth Answer is provided, use it to: (1) understand what information the agent needed to find; (2) identify which responses succeeded vs.\ failed; (3) then analyze: what \textbf{search decisions} (query formulation, result interpretation, search direction) caused the successful responses to find the right information while others did not?\\
        \quad - \textbf{Discriminative Power.} Each rubric must produce actual score variance across the contrasted responses. Focus on observable differences in search behavior. If both sides exhibit the same search pattern, do not create a rubric for it.\\
        \quad - \textbf{Rubric Format.} Each rubric should describe BOTH what a high-scoring response does AND what a low-scoring response does for the same dimension. This helps the judge accurately place each response on the full spectrum.\\
        \quad - \textbf{Balanced Closure Preference.} Do not treat extra verification as an intrinsic virtue. When the retrieved evidence is already sufficient to resolve the key uncertainty, continuing to search, continuing to doubt, or delaying the final answer should be treated as low-scoring behavior rather than ``thoroughness.'' At the same time, do NOT reward premature closure. If important ambiguity, missing constraints, or conflicting evidence remain unresolved, stopping early and answering too soon should also be treated as low-scoring behavior.\\
        \quad - \textbf{Non-Redundancy.} Do not duplicate Existing Rubrics in meaning or scope. If Existing Rubrics already cover the quality differences, return empty lists.\\
        \textbf{Output.}
        Generate 1--3 rubrics total. Fewer high-quality process rubrics are far better than many generic ones.\\
        \texttt{\{ "rubrics": [ \{ "title": "<concise label>", "description": "<what a high-scoring response does>", "counter\_description": "<what a low-scoring response does>" \} ] \}}\\
        ~\\
        \textbf{Inputs.}\\
        \quad 1. \textbf{Question}: The original question the agent is trying to answer.\\
        \quad 2. \textbf{Ground Truth Answer} (if provided): The correct answer --- use to understand what information was needed and which search strategies led to finding it.\\
        \quad 3. \textbf{Contrast Pairs}: One or two higher-target vs.\ lower-target trajectory comparisons.\\
        \quad 4. \textbf{Existing Rubrics}: Previously established rubrics --- do not duplicate these. \\
    \bottomrule
    \end{tabular}
    \caption{\textbf{Contrastive induction prompt.}
    The prompt induces query-local process rubrics from contrastive trajectory pairs.}
    \label{tab:prompt-induction}
\end{table*}

\subsection{Consolidation}
\label{sec:app-prompt-consolidation}

The consolidation prompt is invoked when the candidate pool $\mathcal{D}$ accumulates enough drafts. Existing common rubrics in $\mathcal{P}$ are passed in as a no-duplicate constraint, and the candidate drafts are grouped by query of origin. The full prompt is shown in Table~\ref{tab:prompt-consolidation}.

\begin{table*}[!t]
    \centering
    \fontsize{8pt}{10pt}\selectfont
    \renewcommand{\arraystretch}{0.9}
    \begin{tabular}{p{0.95\linewidth}}
    \toprule
        \rowcolor{gray!20}
        \textbf{Consolidation Prompt} \\
    \midrule
        You are a rubric consolidation expert. Your task is to distill specific, query-level scoring rubrics into \textbf{generalized cross-query standards} that effectively differentiate response quality levels across queries. \\\\

        \textbf{Core Principles}\\
        \quad 1. \textbf{Maximize Differentiation}: Each standard should create meaningful scoring differences between strong and weak responses --- avoid standards that most responses trivially satisfy or all inevitably fail.\\
        \quad 2. \textbf{Align with Sound Reasoning}: Standards must reflect the core elements of correct reasoning --- logical coherence, evidence accuracy, and conclusion reliability --- rather than penalizing minor stylistic differences or irrelevant details.\\
        \quad 3. \textbf{Dual-Sided Standards}: Each standard should describe BOTH the desired behavior (what good responses do) AND the undesired behavior (what weak responses lack) for the same dimension.\\
        \quad 4. \textbf{Balanced Closure}: Do not treat early stopping as a generic virtue. Reward timely closure only when the key uncertainty has already been resolved; do not reward premature closure while ambiguity, missing constraints, or conflicting evidence remain. \\\\

        \textbf{Task}\\
        Identify 1--2 NEW, high-value standards by \textbf{abstracting common patterns} from the candidates below. \\\\

        \textbf{CRITICAL: Do NOT produce any standard that overlaps in meaning with the Existing Common Standards listed below, even if phrased differently.} \\\\

        \textbf{Abstraction Strategy}\\
        \quad - Look for \textbf{recurring quality patterns} across multiple candidates --- both strengths worth rewarding and weaknesses worth penalizing --- then merge them into one general standard with both a DO (desired) and AVOID (undesired) description.\\
        \quad - The description must be \textbf{cross-query general}: applicable to any query, not tied to specific queries.\\
        \quad - Replace question-specific details with general language. Use parenthetical examples ``(e.g., ...)'' to clarify, but the core criterion must be broadly applicable.\\
        \quad\quad - $\times$ Too specific: ``Correctly identifies that hydroxyzine pamoate is for anxiety while hydrochloride is for itching''\\
        \quad\quad - $\checkmark$ Generalized: ``Accurately distinguishes functional or contextual differences between closely related entities (e.g., variant forms, sub-categories, similar-named concepts)''\\
        \quad\quad - $\times$ Too specific: ``Fails to verify the founding year of MIT from the search results''\\
        \quad\quad - $\checkmark$ Generalized: ``Fails to cross-verify key factual claims against retrieved evidence before drawing conclusions''\\
        \quad - Do NOT create standards that penalize trivial issues unrelated to reasoning quality or answer correctness --- stay focused on what genuinely separates strong responses from weak ones.\\
        \quad - If the candidates are already well-covered by Existing Common Standards, return an empty list. \\\\

        \textbf{Existing Common Standards} (already established --- do NOT duplicate these):\\
        \quad - \{existing\_title\_1\}\\
        \quad\quad DO: \{existing\_description\_1\}\\
        \quad\quad AVOID: \{existing\_counter\_description\_1\}\\
        \quad - \{existing\_title\_2\}\\
        \quad\quad DO: \{existing\_description\_2\}\\
        \quad\quad AVOID: \{existing\_counter\_description\_2\}\\
        \quad ... \\\\

        \textbf{New Candidates} (specific rubrics from recent queries):\\
        Query 1: \{query\_1\}\\
        \quad - \{candidate\_title\_1\_1\}\\
        \quad\quad DO: \{candidate\_description\_1\_1\}\\
        \quad\quad AVOID: \{candidate\_counter\_description\_1\_1\}\\
        \quad ...\\
        Query 2: \{query\_2\}\\
        \quad - \{candidate\_title\_2\_1\}\\
        \quad\quad DO: \{candidate\_description\_2\_1\}\\
        \quad\quad AVOID: \{candidate\_counter\_description\_2\_1\}\\
        \quad ... \\\\

        \textbf{Output}\\
        \texttt{\{ "new\_common\_rubrics": [ \{ "title": "...", "description": "<what a high-scoring response does>", "counter\_description": "<what a low-scoring response does>" \} ] \}} \\
    \bottomrule
    \end{tabular}
    \caption{\textbf{Consolidation prompt.}
    The prompt distills recurring patterns from candidate rubrics into reusable common rubrics while avoiding duplicates.}
    \label{tab:prompt-consolidation}
\end{table*}

\subsection{Pairwise Judge}
\label{sec:app-prompt-judge}

Pairwise judging is invoked once per edge in the sparse comparison graph (Section~\ref{sec:reward}). The system prompt fixes the win/tie/loss output format, and the user prompt fills in the question, both responses, and the criterion derived from the active common rubric. Order randomization is applied at the call site by swapping which response receives the A label. The full prompt is shown in Table~\ref{tab:prompt-judge}.

\begin{table*}[!t]
    \centering
    \fontsize{9pt}{11pt}\selectfont
    \renewcommand{\arraystretch}{0.9}
    \begin{tabular}{p{0.95\linewidth}}
    \toprule
        \rowcolor{gray!20}
        \textbf{Pairwise Judge Prompt} \\
    \midrule
        \textbf{System:}\\
        You are an expert judge. You will be given a question (in $<$question$>$$<$/question$>$ tags) and two responses labeled Response A and Response B (in $<$response\_a$>$$<$/response\_a$>$ and $<$response\_b$>$$<$/response\_b$>$ tags). You will also be given a scoring criterion (in $<$criterion$>$$<$/criterion$>$ tags). \\\\

        Your task is to determine which response better satisfies the criterion. Consider the criterion carefully and compare both responses against it. Strictly judge only the given criterion, and do not use any considerations beyond that criterion. \\\\

        Format compliance is part of trajectory quality when it affects the criterion. If one response is malformed, truncated, unparseable, or fails to produce a single clean final answer, that should count against it for criteria involving coherent reasoning, valid tool-use/search-result structure, complete synthesis, or clean finalization. \\\\

        You MUST output JSON in exactly this format: \texttt{\{"winner": "A"\}} or \texttt{\{"winner": "B"\}} or \texttt{\{"winner": "TIE"\}}.\\
        \quad - Output \texttt{\{"winner": "A"\}} if Response A clearly better satisfies the criterion.\\
        \quad - Output \texttt{\{"winner": "B"\}} if Response B clearly better satisfies the criterion.\\
        \quad - Output \texttt{\{"winner": "TIE"\}} if BOTH responses perfectly satisfy the criterion, OR if BOTH fail it entirely, OR if they are virtually indistinguishable in quality regarding this specific criterion. \\\\

        \textbf{User:}\\
        $<$question$>$\{query\}$<$/question$>$\\
        $<$response\_a$>$\{response\_a\}$<$/response\_a$>$\\
        $<$response\_b$>$\{response\_b\}$<$/response\_b$>$\\
        $<$criterion$>$\\
        High-scoring response:\\
        \{rubric\_description\} \\\\

        Low-scoring response:\\
        \{rubric\_counter\_description\}\\
        $<$/criterion$>$ \\
    \bottomrule
    \end{tabular}
    \caption{\textbf{Pairwise judge prompt.}
    The prompt compares two trajectories under one active common rubric and outputs a win, tie, or loss.}
    \label{tab:prompt-judge}
\end{table*}

\subsection{LLM-Judge Accuracy Evaluation}
\label{sec:app-prompt-eval}

Final-answer accuracy on the four benchmarks is reported with an LLM judge that compares the predicted answer against the gold answer. The judge is invoked with the prompt in Table~\ref{tab:prompt-eval} and is configured to output the literal string Correct or Incorrect.

\begin{table*}[!t]
    \centering
    \fontsize{9pt}{11pt}\selectfont
    \renewcommand{\arraystretch}{0.9}
    \begin{tabular}{p{0.95\linewidth}}
    \toprule
        \rowcolor{gray!20}
        \textbf{LLM-Judge Accuracy Evaluation Prompt} \\
    \midrule
        Given a Question and its Golden Answer, verify whether the Predicted Answer is correct. The prediction is correct if it fully aligns with the meaning and key information of the Golden Answer. Respond with ``Correct'' if the prediction is correct and ``Incorrect'' otherwise.\\
        Golden Answer may have multiple options, and matching any one of them is considered correct. \\\\

        Question: \{question\}\\
        Golden Answer: \{labeled\_answer\}\\
        Predicted Answer: \{pred\_answer\} \\
    \bottomrule
    \end{tabular}
    \caption{\textbf{LLM-judge accuracy evaluation prompt.}
    The prompt judges whether a predicted final answer matches the gold answer.}
    \label{tab:prompt-eval}
\end{table*}

\end{document}